# Two-stage residual learning for calibrated uncertainty in solar irradiance forecasts

RR Flanagan[1], E Kasha[1], PW Talbot[2], AS Epiny[2], MR Deinert[1*]

[1] Department of Mechanical Engineering, Colorado School of Mines

[2] Idaho National Laboratory

*Corresponding author. MR Deinert, Department of Mechanical Engineering, Colorado School of Mines.

mdeinert@mines.edu

**Abstract.** Accurate forecasting of renewable generation requires not only point predictions but also reliable estimates of the uncertainty attached to them, particularly at extended horizons. Machine learning models are increasingly used to predict solar output, but they typically report a single value and capture the associated uncertainty poorly. The standard remedy is to train a model to predict the parameters of an assumed probability distribution directly. Here we develop an alternative two-stage method, motivated by synthetic solar signal generation, in which the forecast error is modeled separately from the expected irradiance. A first long short-term memory (LSTM) network predicts the expected global horizontal irradiance, and a second network models the distribution of the forecast error, which is added back to the mean to give both the forecast and its uncertainty. Using historical weather data from the National Solar Radiation Database and a probability integral transform evaluation, we show that explicitly modeling the error brings the predicted standard deviation markedly closer to the historical value, reducing the deviation between the two by 34.6 percent for a standalone LSTM and by 56.5 percent when the LSTM is coupled to Gaussian process regression, the most reliable of the configurations tested. Because the method models the total forecast error directly, it captures the full predictive uncertainty that matters for decisions, rather than only the model contribution that dropout-based approaches estimate.

## Introduction

As the penetration of renewable generators on electrical grids increases, their variability places growing stress on grid operations [1]. Accurate forecasts of renewable generation tell operators how much compensatory generation will be needed [2,3] and support efficient bidding in electricity markets. The field has moved steadily from point forecasts toward probabilistic forecasts that report a full distribution of outcomes [4], a shift that is now well established for solar and photovoltaic power in particular [5]. Solar and wind signals have historically been predicted with statistical methods. The autoregressive moving-average model (ARMA) and its integrated extension (ARIMA) have shown promise for forecasting [6–8]. The inertia in these models makes them well suited to short horizons, but their accuracy falls as the horizon grows. Numerical weather prediction has also been used widely for wind and solar forecasting, though its computational cost can be significant [9].

Machine learning offers another route to predicting solar output [10,11]. A least absolute shrinkage and selection operator (LASSO) approach has been used to forecast solar power generation from five-minute weather data in Massachusetts [12]. Several reviews report that the best machine learning method often depends on the geographic area [13,14]. Others combine statistical and machine learning techniques. Xie and colleagues combined variational mode decomposition, deep belief networks, and ARMA to predict solar signals, and reported improvements over each method used alone [15]. Long short-term memory (LSTM) networks have been used for hour-ahead and day-ahead solar irradiance prediction and can outperform feedforward neural networks [16], and an LSTM trained partly on non-local data has been used to predict the next day of hourly irradiance from local weather forecasts [17]. Reviews of deep learning for solar forecasting find that hybrid models such as the convolutional LSTM can outperform a standalone LSTM, although the strongest results in those reviews came from notably dry, clear-sky locations [18].

Neural networks have also been applied to other natural signals. A discrete wavelet transform combined with an LSTM has outperformed wavelet combinations with regression and backpropagation networks for short-term wind power [19]. An LSTM has also outperformed a moving-average approach and a multilayer perceptron on mean square error in turbine SCADA data [20]. Whatever the forecasting approach, the uncertainty attached to a prediction is both essential and too often missing. This is a recognized problem for machine learning models generally [21]. One way to estimate uncertainty is to assume the modeled quantity is Gaussian and to train a network to predict the mean and standard deviation of that Gaussian for a given set of weather inputs [22], which gives both an expected value and a distribution. A network can also be trained to output the parameters of a richer conditional distribution, as in a mixture density network [23]. This family of methods has been used widely, including for the estimation of nuclear cross sections [24]. Gaussian process models provide another route to uncertainty [25]. Gaussian process regression has been used to predict hour-ahead solar panel output with attached confidence intervals [26]. For solar power, coupling an LSTM to a Gaussian process reduced the uncertainty in the result relative to an LSTM alone [27], and the same pairing has been applied to short-term wind speed [28]. Gaussian process models are, however, computationally expensive and scale poorly.

Despite these advances, current methods remain limited in their ability to capture uncertainty fully. Traditional machine learning often fails to account for both aleatoric uncertainty, which arises from irreducible randomness, and epistemic uncertainty, which reflects gaps in knowledge and is in principle reducible [21]. Most methods address aleatoric uncertainty by modeling noise in the data but rarely capture the model's own ignorance, so predictions can appear overconfident where data are sparse or unrepresentative [21]. Several approaches target the epistemic part directly. Monte Carlo dropout deactivates connections at random during both training and inference, producing an ensemble of stochastic forward passes whose spread estimates uncertainty [29]. Deep ensembles train several networks from different initializations and combine their predictions [30]. A single network can also be trained to output both the data noise and a measure of model uncertainty at once [31], or to return the parameters of an evidential distribution from one forward pass [32]. Each of these mainly captures one component, and the spread they report is a proxy rather than a direct measure of the total error.

Isolating the uncertainty in a prediction would allow the aleatoric and epistemic contributions to be measured together. This is difficult for time series forecasts of probability distributions, because there is usually no ground truth distribution to compare against, only single realized outcomes. Earlier work on synthetic solar signals showed that isolating and modeling the noise component, separately from the clear-sky irradiance, improves signal generation [33]. Building on that result, we combine noise modeling with machine learning for solar forecasting and compare two approaches. In the first, an LSTM predicts the parameters of a modified Gaussian for the irradiance directly. In the second, the expected value is subtracted from the ground truth to build a dataset of forecast errors, and a second LSTM predicts the distribution of that error, which is added back to the expected value to give both the forecast and its distribution. Both approaches use historical weather data from the National Solar Radiation Database [34]. The difference is that the two-stage method gives an explicit estimate of the error distribution, whereas the single-stage method gives only an implicit one. We evaluate both with the probability integral transform [35], and we extend the comparison to configurations that couple each method to Gaussian process regression. A standalone LSTM was chosen for the core comparison so that the benefit of the two-stage method could be isolated, though hybrid architectures such as the convolutional LSTM have shown promise for solar forecasting [18].

## Methods

### *Data*

Ground truth data were collected from the National Solar Radiation Database (NSRDB) [34] for Austin, Texas and Seattle, Washington, together with the Department of Energy Open Energy Data Initiative. Each record contains wind speed, wind direction, solar strength, dew point, temperature, humidity, air pressure, time of day, and time of year. An additional field representing cloud cover was derived by subtracting the historical value from the calculated clear-sky maximum. The historical data are used to make hour-ahead and day-ahead predictions.

***Data processing***

Records for Austin and Seattle were drawn at 30-minute resolution. Weather conditions from 24 surrounding regions, upstream and downstream of each target site, were included to provide pseudo-forecast information. All fields were scaled to the interval from zero to one to improve the performance of the network.

Two fields were derived from the raw data: normalized global horizontal irradiance (GHI) and estimated cloud cover. Normalized GHI is the GHI divided by the maximum solar irradiance at the same time, and is the prediction target. Estimated cloud cover is the clear-sky maximum minus the historical GHI, normalized by the clear-sky maximum, and represents the fraction of sunlight blocked by clouds.

The calculated maximum solar irradiance is used to identify intervals when irradiance is necessarily zero. These nighttime intervals are excluded from the root mean square error, the coefficient of determination, and the error calculations, because irradiance is known to be zero at night and a network is not required to predict it.

***Parameter prediction method***

The parameter prediction method uses an LSTM network to predict the parameters of a modified Gaussian, Equation (1):

$$f(t+\Delta t) = \frac{1}{\sqrt{2\pi c}} e^{-\frac{(x-a)^2}{2b^2}} \tag{1}$$

Here f is the probability density of the GHI at time t+Δt, and the network predicts the parameters a, b, and c, from which the probability of a given value of GHI(t+Δt) follows. Figure 1 (PPM) shows the process. Equation (1) has the form of a Gaussian, but the three parameters are allowed to vary independently, which makes the function more general than a standard Gaussian, whose normalization is fixed by its width. The parameters are obtained by training the network to maximize the likelihood of the observed irradiance under Equation (1), equivalently by minimizing the negative log-likelihood, Equation (2), averaged over the training data:

$$L = \frac{1}{N}\sum_{i=1}^{N}\left[\frac{(x_i-a_i)^2}{2{b_i}^2} + \frac{1}{2}\ ln(2\pi c_i)\right] \tag{2}$$

***Augmented error method***

The parameter prediction method gives a prediction of the uncertainty. The uncertainty can also be estimated by comparing the predicted value with the historical value. For this, the predicted value is subtracted from the historical GHI, and the difference is the error of that prediction.

The forecast errors are generated with a leave-one-year-out procedure, so that no year is predicted by a model that was trained on it. For each year in the twenty-year record from 2002 to 2021, a parameter prediction network is trained on the remaining nineteen years and used to predict that year. The expected value of each prediction is subtracted from the corresponding ground truth to give the forecast error at 30-minute resolution across the full record. Figure 1 (AEM) shows the process, where f in Equation (1) now represents the error at time t+Δt rather than the GHI.

A second LSTM is then trained on the historical weather data from 2002 to 2021 together with these errors, and is applied over the test period from 2022 to 2024 to predict an error distribution for each 30-minute interval. Adding each error distribution back to the expected GHI gives the forecast and its distribution. The test period yields approximately 16,000 predicted distributions per configuration, which are analyzed below. The specifications of each network are given in the Supplementary Information. This two-stage structure is a form of residual learning where the second network models the residual of the first-stage prediction rather than the quantity of interest directly.

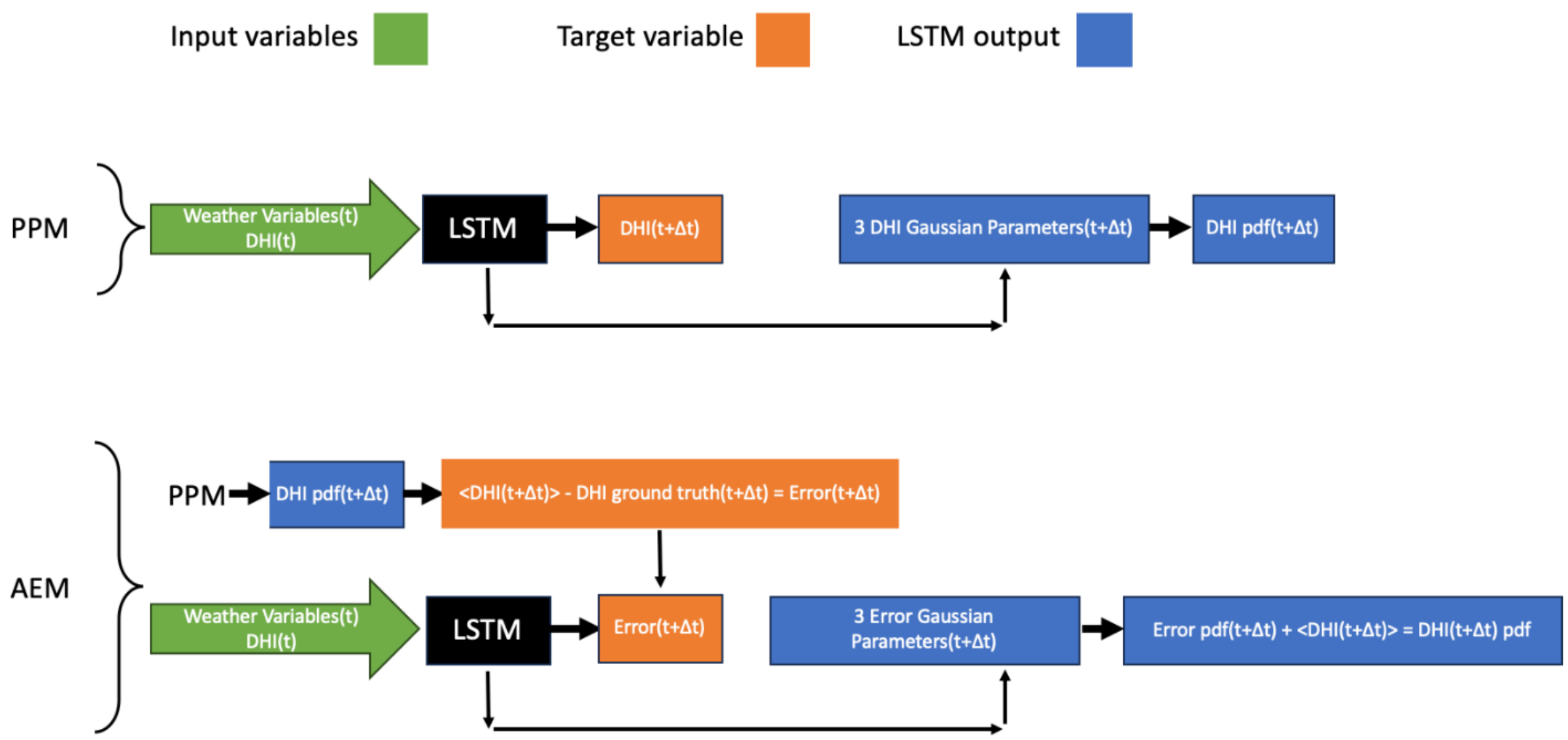

**Figure 1.** Schematic for training the parameter prediction method (PPM) and the augmented error method (AEM). Historical ground truth data at time t, for both GHI and weather, are used to train the PPM to predict the GHI distribution at time t+Δt. The training data contain both the t and t+Δt values, which are used to generate a semi-synthetic error dataset for time t+Δt. The AEM uses this dataset to predict the GHI distribution at t+Δt. Once trained, both models require only weather and GHI values at time t to predict the GHI distribution at t+Δt.

***LSTM setup***

Long short-term memory networks process sequences rather than single data points, which suits time series because earlier values can influence the prediction of later ones. An LSTM cell has three gates that manage its information: an input gate selects which information is updated in the cell, a forget gate determines which information is kept and which is discarded, and an output gate returns the cell result [36,37]. The networks were built with the Keras interface to TensorFlow [38]. The network hyperparameters, including the number of epochs, layers, neurons, batch size, and dropout rate, were selected with the Optuna optimization framework [39]. The search ranges and the selected values are reported in the Supplementary Information.

***Gaussian process regression***

A combined LSTM and Gaussian process regression model was also tested with the same procedures. Gaussian process models are an established route to uncertainty estimation in machine learning [25–28]. Here the LSTM was trained to predict the GHI rather than the parameters of a distribution, and the Gaussian process supplied the distribution. The combined model was also run through the augmented error procedure, in which the LSTM again predicted an error rather than a distribution of error. As with the parameter prediction network, residuals from 2002 to 2021 were used to build this model.

***Evaluating the predicted distributions***

If a distribution correctly represents the uncertainty in a prediction, then ground truth values fall within it at the frequencies the distribution implies [35]. This property is the basis of the probability integral transform, and uniformity of the transformed values is the standard diagnostic for a calibrated probabilistic forecast. We use it to compare the methods. We also summarize accuracy with the root mean square error of the point forecast, the continuous ranked probability score [40], and the empirical coverage of the central prediction intervals at the 95, 90, and 50 percent levels. We do not report the logarithmic score. It evaluates the predicted probability density at the realized value, and the modified Gaussian of Equation (1), whose parameters are independent, does not integrate to one, so its density is fixed only up to a constant. The probability integral transform and the continuous ranked probability score depend only on the cumulative distribution, in which that constant cancels, and so remain well defined.

Each predicted distribution is divided into 1000 bins of equal probability, each holding one part in a thousand of the total probability. The bin edges are found by integrating the distribution until the cumulative probability reaches each successive multiple of one part in a thousand. The ground truth value for that time step is placed in the bin into which it falls. Counts are accumulated across all predicted distributions in the test period, which gives a histogram over the 1000 probability bins. Figure 2 shows the process.

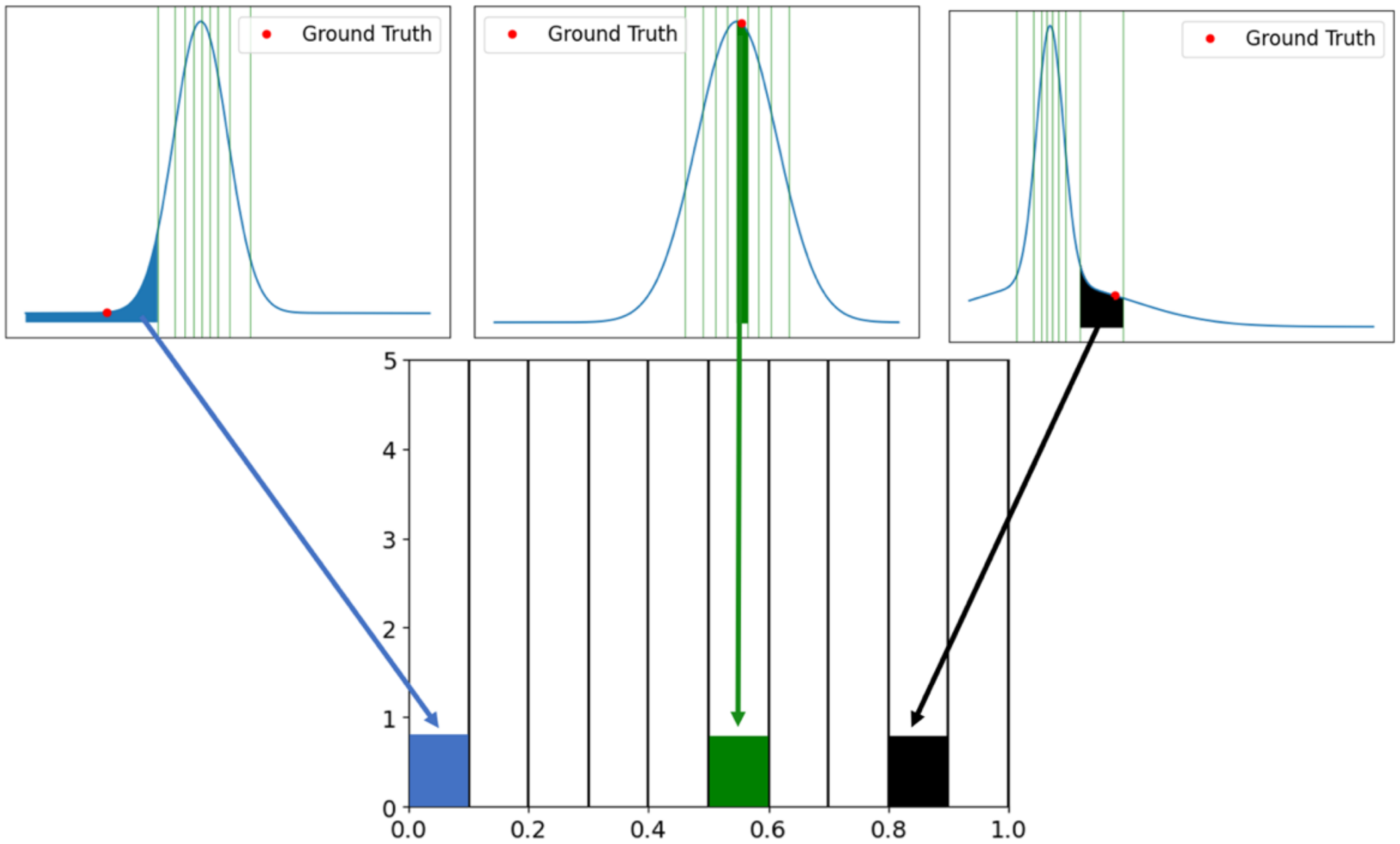

**Figure 2.** The probability integral transform. The model produces a set of values for a, b, and c in Equation (1) at each time point. Because there is no ground truth distribution to compare against, each predicted distribution is divided into bins of equal cumulative probability, and the ground truth value is placed in the bin into which it falls. If the predicted distributions are correct, the binned values form a uniform histogram.

If the distributions are correct, the histogram is uniform, with each bin holding the total number of values divided by the number of bins. With approximately 16,000 values this is about 16 per bin. Departures from a uniform histogram indicate that the predicted distributions do not match the data.

We test each histogram against the uniform expectation with a chi-square goodness-of-fit test, Equation (3), where O_i is the count in bin i and E is the expected count, equal to the number of values divided by 1000. The test has 999 degrees of freedom, one fewer than the number of bins:

$$\chi^2 = \sum_{i=1}^{1000} \frac{(O_i - E)^2}{E}, \; E = \frac{N}{1000} \tag{3}$$

The chi-square test is the appropriate choice here. Because every ground truth value falls into exactly one bin, the bin counts sum to the number of predicted values and their mean is fixed at about 16 for every method, regardless of calibration. The information that distinguishes a calibrated forecast from a miscalibrated one is therefore carried entirely by the spread of the counts about this value, which is exactly the quantity the chi-square statistic measures.

**Results and discussion**

Many forecasting methods, including those built on deep learning, focus on point predictions and do not report the associated probability distribution. Methods that do address uncertainty often assume a single distribution for the forecast error, usually Gaussian [22]. Forecast errors can, however, show skew, multiple peaks, and heavy tails that a single assumed distribution does not capture [28]. Capturing both the aleatoric and epistemic parts of the uncertainty is also difficult, particularly the latter [21]. The method developed here addresses these issues by making an explicit estimate of the uncertainty in a predicted irradiance and then using an LSTM to choose the parameters of a modified Gaussian that fits it.

Table 1 reports the average predicted standard deviation over a year and the average historical values for Austin and Seattle, for the parameter prediction method, the augmented error method, the LSTM with Gaussian process regression, and the LSTM with Gaussian process regression in its augmented error form, at both horizons. Nighttime intervals, where the signal is known to be zero, are excluded, since including them would inflate the apparent accuracy. The parameter prediction and augmented error methods share the same base LSTM and differ in how the forecast error is modeled. The historical standard deviation at a single time step does not represent the true value at that step and is shown only for reference. In every case the augmented error method gives an average predicted standard deviation closer to the historical value than its single-stage counterpart. Averaged across the cases, the augmented error method reduces the deviation between the predicted and historical standard deviation by 34.6 percent for the standalone LSTM and by 56.5 percent for the LSTM with Gaussian process regression. The latter is closest to the historical standard deviation in every case examined.

| **Method** | **Austin predicted standard deviation** (W/m²) | **Austin historical standard deviation** (W/m²) | **Seattle predicted standard deviation** (W/m²) | **Seattle historical standard deviation** (W/m²) |
|---|---|---|---|---|
| **One-hour prediction** | | | | |
| Parameter prediction method | 30.2 | 34.4 | 41.3 | 45.1 |
| Augmented error method | 36.9 | | 47.6 | |
| LSTM with | 36.2 | | 47.2 | |
| Gaussian process regression | | | | |
| LSTM with Gaussian process regression, augmented error | 35.4 | | 44.3 | |
| **24-hour prediction** | | | | |
| Parameter prediction method | 29.4 | 34.4 | 39.9 | 45.1 |
| Augmented error method | 31.2 | | 41.4 | |
| LSTM with Gaussian process regression | 37.6 | | 49.5 | |
| LSTM with Gaussian process regression, augmented error | 32.4 | | 43.9 | |

Table 1. Average predicted standard deviation over a year and average historical standard deviation for Austin and Seattle, for the parameter prediction method, the augmented error method, the LSTM with Gaussian process regression, and its augmented error form, at the one-hour and 24-hour horizons. The augmented error method is closer to the historical standard deviation in every case. The historical standard deviation is shown once per horizon as a reference and does not represent the true standard deviation at a single time step.

Figure 3 shows the probability integral transform histograms for the hour-ahead GHI distributions from the parameter prediction and augmented error methods. The top two panels are for Austin, with the parameter prediction method on the left and the augmented error method on the right, and the bottom two panels are the same for Seattle. A perfectly calibrated set of distributions would place about 16 values in each bin with no systematic departure, and the further a histogram departs from this, the less accurate the predicted distributions. Figure 4 gives the corresponding results for day-ahead predictions. Figures 5 and 6 show the results for the LSTM with Gaussian process regression and its augmented error form. The augmented error variants give visibly flatter histograms, and the augmented error form of the LSTM with Gaussian process regression is the flattest of the four configurations.

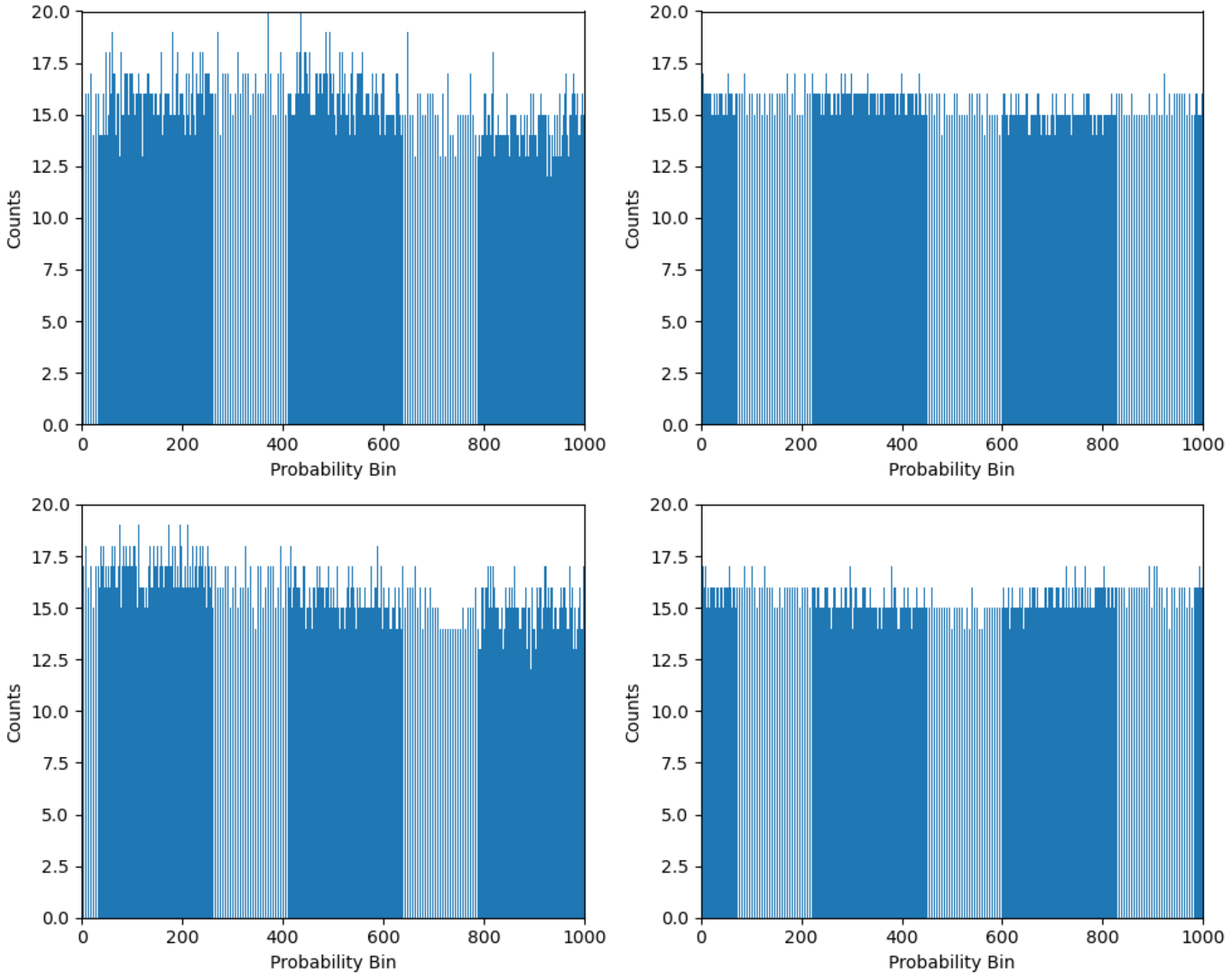

**Figure 3.** Probability integral transform histograms of the hour-ahead GHI distributions from the parameter prediction and augmented error methods. The top panels are for Austin, Texas, with the parameter prediction method on the left and the augmented error method on the right. The bottom panels are the same for Seattle, Washington.

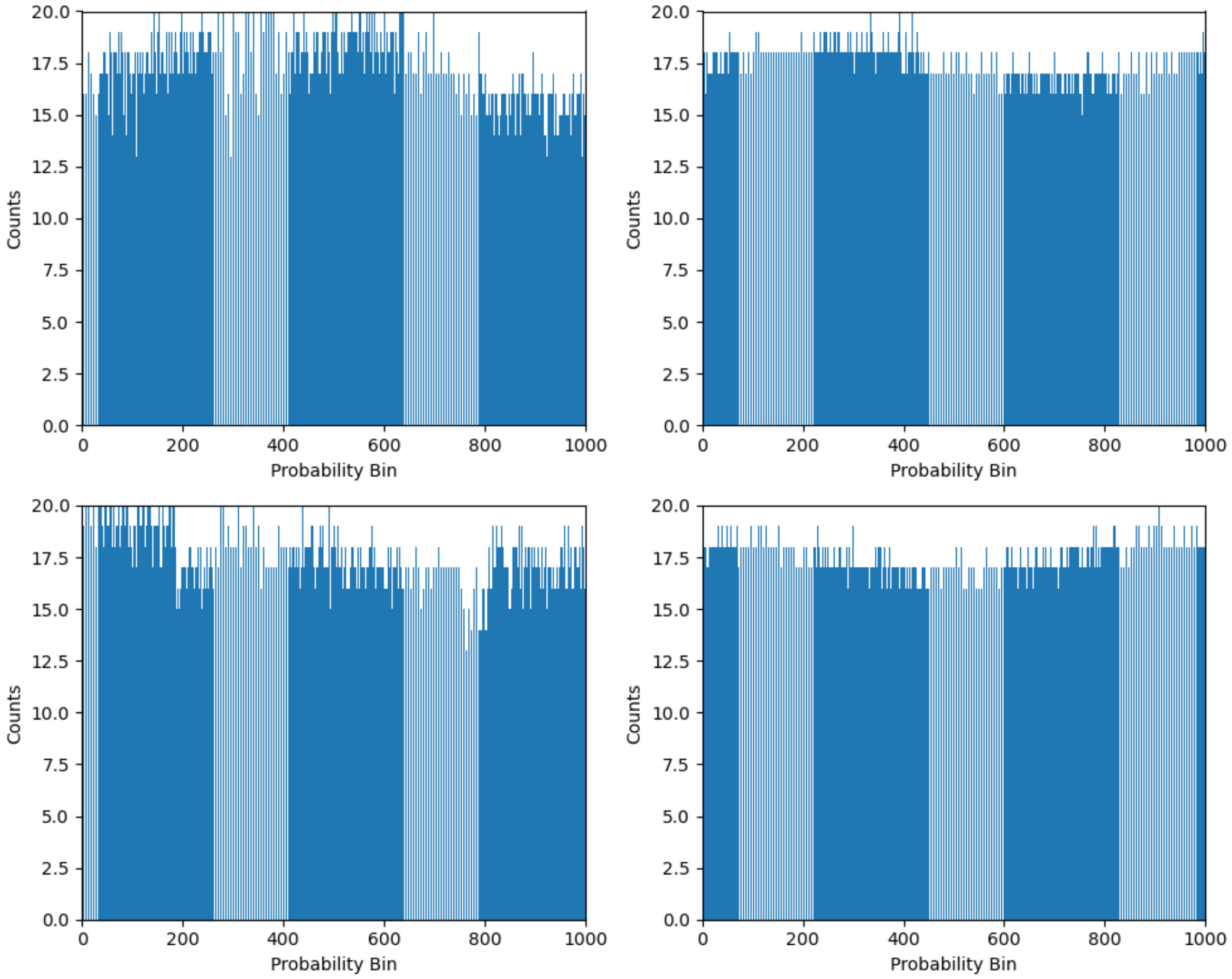

**Figure 4.** Probability integral transform histograms of the day-ahead GHI distributions from the parameter prediction and augmented error methods. The top panels are for Austin, Texas, with the parameter prediction method on the left and the augmented error method on the right. The bottom panels are the same for Seattle, Washington.

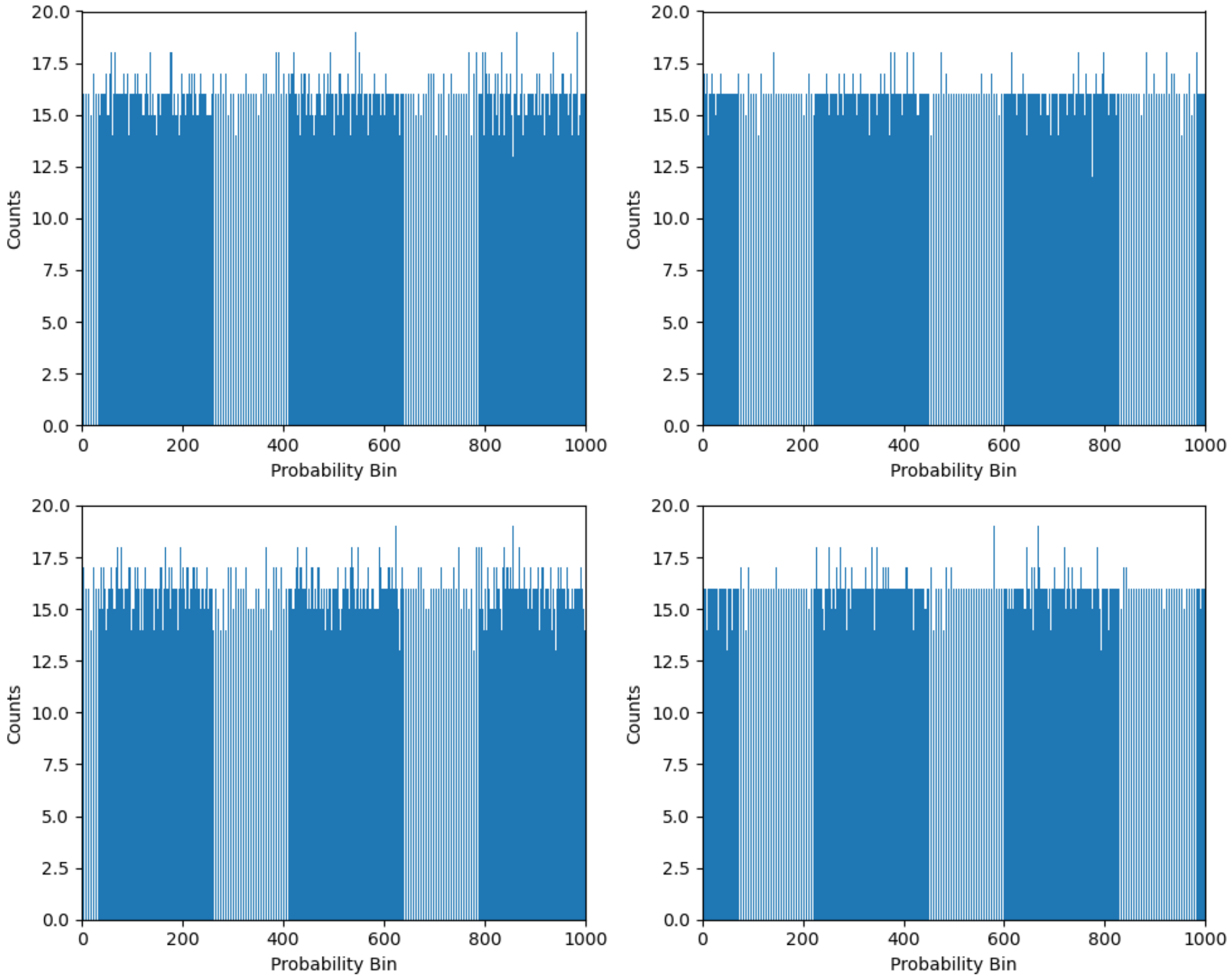

**Figure 5.** Probability integral transform histograms of the hour-ahead GHI distributions from the LSTM with Gaussian process regression. The top panels are for Austin, Texas, with the standalone model on the left and its augmented error form on the right. The bottom panels are the same for Seattle, Washington.

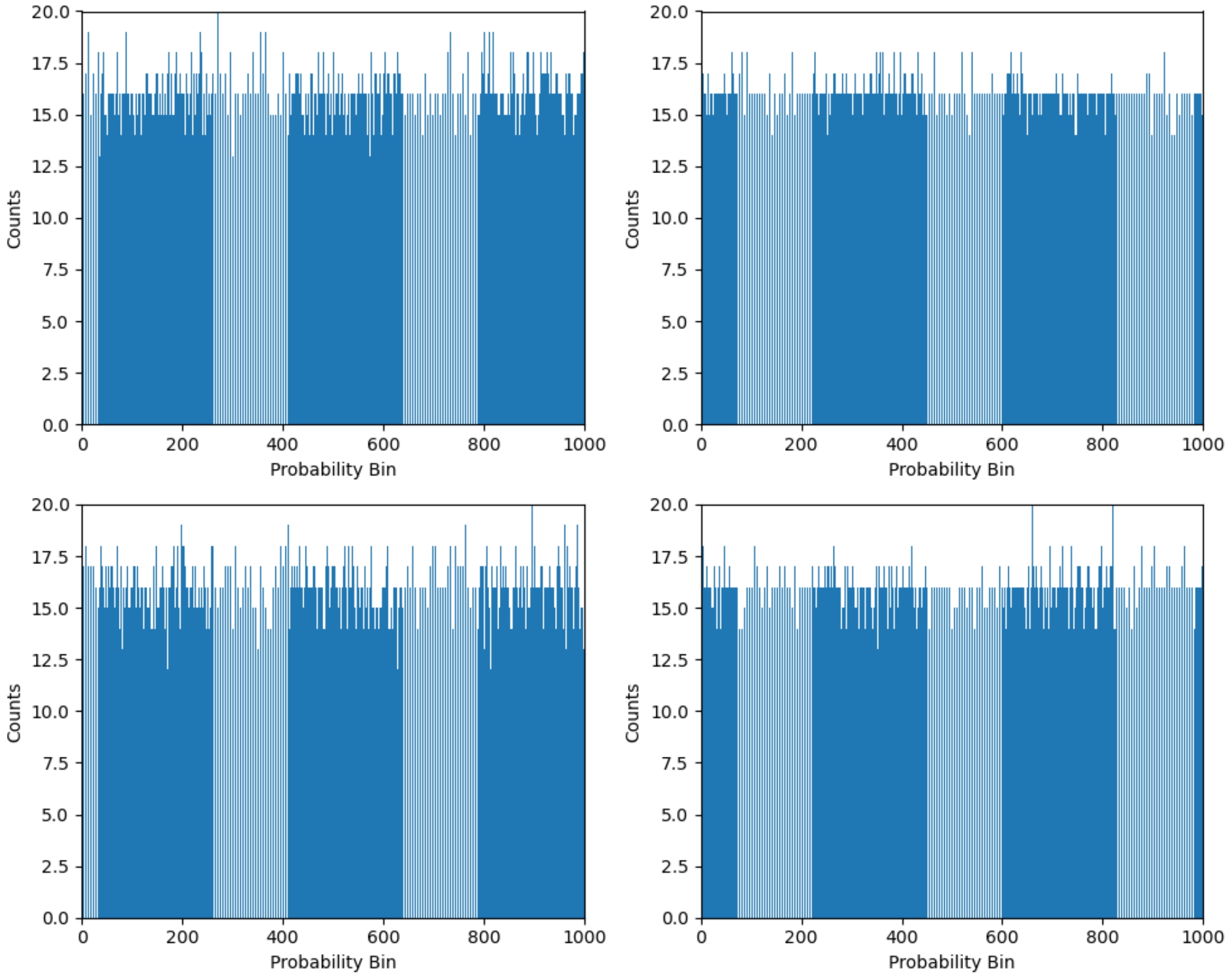

**Figure 6.** Probability integral transform histograms of the day-ahead GHI distributions from the LSTM with Gaussian process regression. The top panels are for Austin, Texas, with the standalone model on the left and its augmented error form on the right. The bottom panels are the same for Seattle, Washington.

We compared each histogram with the uniform expectation using a chi-square goodness-of-fit test with 999 degrees of freedom. The results are given in Table 2. The histograms from the parameter prediction method and from the standalone LSTM with Gaussian process regression differ significantly from the uniform expectation, with p-values below 0.05, so these methods do not reproduce the expected distribution. Every configuration that uses the augmented error method has a p-value above 0.05, and so is not distinguishable from the uniform expectation. Taken together with Table 1, the augmented error form of the LSTM with Gaussian process regression is the most reliable configuration: it passes the calibration test and its predicted standard deviation is closest to the historical standard deviation in every case.

| Method | Austin | Seattle |
|---|---|---|
| Parameter prediction method, 1 h | $1.1\times10^{-8}$ | $3.0\times10^{-9}$ |
| Parameter prediction method, 24 h | $4.6\times10^{-11}$ | $1.5\times10^{-13}$ |
| Augmented error method, 1 h | 0.47 | 0.38 |
| Augmented error method, 24 h | 0.18 | 0.47 |
| LSTM with Gaussian process regression, 1 h | $2.2\times10^{-3}$ | $2.0\times10^{-4}$ |
| LSTM with Gaussian process regression, 24 h | $9.0\times10^{-4}$ | $1.1\times10^{-4}$ |
| LSTM with Gaussian process regression, augmented error, 1 h | 0.39 | 0.25 |
| LSTM with Gaussian process regression, augmented error, 24 h | 0.45 | 0.21 |

**Table 2.** Chi-square goodness-of-fit p-values comparing each probability integral transform histogram with the uniform expectation, for each city, method, and horizon, with 999 degrees of freedom. A p-value at or below 0.05 indicates a histogram that differs significantly from the uniform expectation.

The probability integral transform evaluates the entire predicted distribution against the realized values at every probability level. A method that passes it has reproduced the conditional distribution of the outcome, not merely a summary statistic of it. A flat histogram is, strictly, a necessary rather than a sufficient condition, because opposing miscalibrations could in principle cancel, but such cancellation would have to hold across the entire record at every probability level, which a single consistently trained model does not produce and which the clear rejection of the parameter prediction and standalone Gaussian process configurations shows the test still detects. A single summary score such as the continuous ranked probability score compresses calibration and sharpness into one value, and a model can attain a competitive score while remaining miscalibrated [40]. Because operational decisions depend on the probability assigned to each possible outcome, we treat the probability integral transform as the primary measure of calibration and report the summary scores below as a complement.

Calibration describes the shape of the predicted distribution. The accuracy of the forecasts can also be summarized with scores that compare the distribution, or its center, with the realized values. Table 3 reports the root mean square error of the point forecast and the continuous ranked probability score, and Table 4 reports the empirical coverage of the central prediction intervals at the 95, 90, and 50 percent levels, for both horizons and both sites.

The augmented error method lowers both the root mean square error and the continuous ranked probability score relative to its base in every case. Averaged over the cases, the continuous ranked probability score falls by about 33 percent for the standalone LSTM and by about 22 percent for the LSTM with Gaussian process regression. The augmented error form of the LSTM with Gaussian process regression gives the lowest root mean square error and the lowest continuous ranked probability score in most cases, which agrees with its being the best calibrated. As an external baseline we also evaluate Monte Carlo dropout applied to the base LSTM, with the predictive distribution taken from the spread of 500 stochastic forward passes at inference. It improves on the single-stage parameter prediction method but does not reach either augmented error method on the root mean square error or the continuous ranked probability score in any case, which leaves the augmented error form of the LSTM with Gaussian process regression the best overall. The central prediction intervals are close to their nominal levels for the four methods and the Monte Carlo dropout baseline alike, and the configurations that use Gaussian process regression are marginally closer to the nominal 95 percent coverage than the others. Because the two methods share the same base LSTM, the improvement in root mean square error is attributable to second stage. Here, the predicted error distribution can have a nonzero expected value, which acts as an effective bias correction to the first-stage point forecast.

| **Method** | **Austin RMSE ($W/m^2$)** | **Austin CRPS ($W/m^2$)** | **Seattle RMSE ($W/m^2$)** | **Seattle CRPS ($W/m^2$)** |
|---|---|---|---|---|
| **One-hour prediction** | | | | |
| Parameter prediction method | 33.6 | 16.9 | 29.0 | 14.2 |
| Augmented error method | 22.8 | 11.5 | 20.7 | 10.1 |
| LSTM with Gaussian process regression | 27.8 | 14.0 | 24.2 | 11.8 |
| LSTM with Gaussian process regression, augmented error | 21.8 | 11.0 | 18.8 | 9.2 |
| Monte Carlo dropout | 28.3 | 14.3 | 25.9 | 12.6 |
| **24-hour prediction** | | | | |
| Parameter prediction method | 36.5 | 18.3 | 33.0 | 16.0 |
| Augmented error method | 23.4 | 11.8 | 21.5 | 10.4 |
| LSTM with Gaussian process regression | 28.4 | 14.3 | 25.4 | 12.3 |
| LSTM with Gaussian process regression, augmented error | 23.7 | 12.0 | 18.0 | 8.8 |
| Monte Carlo dropout | 33.0 | 16.6 | 29.1 | 14.1 |

**Table 3.** Root mean square error and continuous ranked probability score, both in watts per square meter, for the four methods and a Monte Carlo dropout baseline at the one-hour and 24-hour horizons for Austin and Seattle. Lower values are better. The augmented error method improves both over its base in every case, and both augmented methods improve on the Monte Carlo dropout baseline.

| Method | Austin 95% | Austin 90% | Austin 50% | Seattle 95% | Seattle 90% | Seattle 50% |
|---|---|---|---|---|---|---|
| **One-hour prediction** | | | | | | |
| Parameter prediction method | 92.6 | 86.6 | 46.4 | 92.7 | 86.7 | 46.2 |
| Augmented error method | 92.5 | 86.4 | 46.1 | 91.7 | 85.2 | 44.7 |
| LSTM with Gaussian process regression | 93.5 | 87.7 | 47.1 | 93.2 | 87.5 | 46.9 |
| LSTM with Gaussian process | 93.1 | 87.4 | 46.9 | 93.2 | 87.4 | 46.9 |
| regression, augmented error | | | | | | |
| Monte Carlo dropout | 92.6 | 86.6 | 46.4 | 91.7 | 85.3 | 45.5 |
| **24-hour prediction** | | | | | | |
| Parameter prediction method | 90.6 | 84.5 | 44.2 | 92.1 | 86.1 | 45.6 |
| Augmented error method | 91.9 | 85.6 | 45.2 | 90.8 | 84.0 | 43.8 |
| LSTM with Gaussian process regression | 92.2 | 86.3 | 45.7 | 92.2 | 86.3 | 45.8 |
| LSTM with Gaussian process regression, augmented error | 93.5 | 87.9 | 47.1 | 94.0 | 88.5 | 47.9 |
| Monte Carlo dropout | 89.1 | 82.5 | 42.2 | 90.6 | 83.8 | 44.0 |

**Table 4.** Empirical coverage, in percent, of the central prediction intervals at the 95, 90, and 50 percent nominal levels, for the four methods and a Monte Carlo dropout baseline at the one-hour and 24-hour horizons for Austin and Seattle. Values close to the nominal level indicate well-sized intervals.

There are several limitations to the current work. An LSTM can in principle shift historical data to reproduce a signal. To ensure that a predicted value is not a copy of the historical signal, leading or lagging, a correlation test was run comparing the predicted signal against the historical signal shifted by up to 100 time steps in each direction, and any near-perfect match was inspected visually to confirm that no time shifting had occurred. Many physical weather variables could not be included here because of the limited public data, among them the convective available potential energy and atmospheric temperature gradients that underlie deterministic weather models. Including such variables in the training set could improve the results. Finally, the comparison here uses a standalone LSTM by design, so that the benefit of the two-stage method can be isolated. Other probabilistic approaches, such as boosting-based distributional regression [41], could be evaluated with the same probability integral transform in future work.

**Conclusions**

This work introduces a two-stage augmented error method for modeling the probability distributions of solar irradiance forecasts and evaluates it against a conventional single-stage approach and against configurations coupled to Gaussian process regression. The method addresses several persistent gaps in machine learning forecasting. Conventional models, including LSTMs, tend to report point forecasts without the associated distribution, which limits their use where decisions depend on risk, such as grid operation and market bidding. Dedicating a second network to the forecast error provides that distribution explicitly. Where earlier work assumes a single fixed error distribution and so misses skew, heavy tails, and multiple peaks, the augmented error method fits a modified Gaussian to the explicitly modeled error, and could be extended to other distributions. By modeling the total forecast error directly, it captures the full predictive uncertainty that decisions depend on. Methods built on Monte Carlo dropout estimate only the model contribution, and methods that assume a single noise distribution estimate only the data contribution. The augmented error method requires neither component to be isolated, because the total error is what a forecast user needs.

Evaluating uncertainty is itself difficult, because there is no ground truth distribution to compare against. The probability integral transform addresses this by testing whether ground truth values fall within the predicted distributions at the expected frequencies. Under a chi-square goodness-of-fit test, the augmented error distributions are not distinguishable from the uniform expectation, while the single-stage parameter prediction and the standalone Gaussian process configurations are. Coupling the augmented error method to Gaussian process regression gives the best result overall, since it is the only configuration whose predicted standard deviation is closest to the historical standard deviation in every case while also passing the calibration test.

Across the cases studied, the augmented error method reduces the deviation between the predicted and historical standard deviation by 34.6 percent for the standalone LSTM and by 56.5 percent for the LSTM with Gaussian process regression, and it produces distributions that better reflect the range of solar irradiance outcomes. By making the uncertainty explicit and testing it directly, the approach improves the reliability of machine learning forecasts for renewable energy integration and for other settings where the distribution of outcomes matters.

**Data availability**

The training and test data will be made available upon request.

**Supplementary Information**

*Two-stage residual learning for calibrated uncertainty in solar irradiance forecasts*

## Machine learning algorithm operation

The hyperparameters of each network were selected with the Optuna optimization framework, as described in the main text. The parameters swept were the number of epochs, the batch size, the number of neurons, and the dropout rate, which sets the fraction of neuron connections switched off. The dropout rate was held constant across layers when a network had more than one layer, and the number of layers was optimized by running the tuner for each candidate number of layers. Table S1 gives the range searched for each parameter.

| Parameter | Low | High |
| --- | --- | --- |
| Epochs | 1 | 2000 |
| Layers | 1 | 4 |
| Batch size | 1 | 500 |
| Neurons | 1 | 500 |
| Dropout rate | 0.01 | 0.3 |

**Table S1.** Range searched for each hyperparameter in the Optuna sweep.

After the sweep, the configuration that gave the smallest root mean square error for each field was selected. These configurations are listed in Tables S2 to S5.

| Field | Epochs | Layers | Batch size | Neurons | Dropout rate |
| --- | --- | --- | --- | --- | --- |
| 1 hour, Austin | 1266 | 1 | 47 | 13 | 0.035 |
| 1 hour, Seattle | 1399 | 1 | 49 | 14 | 0.044 |
| 24 hour, | 1874 | 1 | 46 | 13 | 0.023 |
| Austin | | | | | |
| 24 hour, Seattle | 1801 | 1 | 46 | 20 | 0.021 |

**Table S2.** Optimized network hyperparameters for the parameter prediction method.

| Field | Epochs | Layers | Batch size | Neurons | Dropout rate |
|---|---|---|---|---|---|
| 1 hour, Austin | 1681 | 1 | 27 | 9 | 0.033 |
| 1 hour, Seattle | 1722 | 1 | 24 | 10 | 0.023 |
| 24 hour, Austin | 1337 | 1 | 22 | 10 | 0.041 |
| 24 hour, Seattle | 1201 | 1 | 21 | 12 | 0.039 |

**Table S3.** Optimized network hyperparameters for the augmented error method.

| Field | Epochs | Layers | Batch size | Neurons | Dropout rate |
|---|---|---|---|---|---|
| 1 hour, Austin | 854 | 2 | 24 | 35 | 0.022 |
| 1 hour, Seattle | 777 | 2 | 23 | 33 | 0.036 |
| 24 hour, Austin | 916 | 2 | 18 | 24 | 0.038 |
| 24 hour, Seattle | 988 | 2 | 19 | 19 | 0.041 |

**Table S4.** Optimized network hyperparameters for the LSTM with Gaussian process regression.

| Field | Epochs | Layers | Batch size | Neurons | Dropout rate |
|---|---|---|---|---|---|
| 1 hour, Austin | 335 | 1 | 18 | 6 | 0.043 |
| 1 hour, Seattle | 197 | 1 | 17 | 4 | 0.051 |
| 24 hour, Austin | 118 | 1 | 28 | 16 | 0.062 |
| 24 hour, Seattle | 234 | 1 | 21 | 7 | 0.098 |

**Table S5.** Optimized network hyperparameters for the augmented error form of the LSTM with Gaussian process regression.